# Language-Driven Coordination and Learning in Multi-Agent Simulation Environments

Zhengyang Li, *Member, IEEE*, Sawyer Campos, Nana Wang

***Abstract*— This paper introduces LLM-MARL, a unified framework that incorporates large language models (LLMs) into multi-agent reinforcement learning (MARL) to enhance coordination, communication, and generalization in simulated game environments. The framework features three modular components of Coordinator, Communicator, and Memory, which dynamically generate subgoals, facilitate symbolic inter-agent messaging, and support episodic recall. Training combines PPO with a language-conditioned loss and LLM query gating. LLM-MARL is evaluated in Google Research Football, MAgent Battle, and StarCraft II. Results show consistent improvements over MAPPO and QMIX in win rate, coordination score, and zero-shot generalization. Ablation studies demonstrate that subgoal generation and language-based messaging each contribute significantly to performance gains. Qualitative analysis reveals emergent behaviors such as role specialization and communication-driven tactics. By bridging language modeling and policy learning, this work contributes to the design of intelligent, cooperative agents in interactive simulations. It offers a path forward for leveraging LLMs in multi-agent systems used for training, games, and human-AI collaboration.**

## I. INTRODUCTION

Multi-agent reinforcement learning (MARL) has become a cornerstone for enabling intelligent cooperation and competition in complex, dynamic environments. From autonomous vehicle coordination to real-time strategy games and swarm robotics, MARL empowers systems of agents to make decisions that maximize collective or individual rewards through interaction and learning. Despite numerous breakthroughs, including the development of algorithms like QMIX, MAPPO, and MADDPG, MARL systems continue to face major obstacles in communication efficiency, task generalization, strategic coordination, and long-term memory management.

Traditional MARL approaches rely heavily on structured state representations, end-to-end learned policies, and domain-specific coordination rules, which often fail to scale to novel or language-driven tasks. One of the core limitations lies in agents' inability to process symbolic instructions, reuse high-level knowledge, or effectively communicate in naturalistic ways. These constraints result in brittle agents that underperform in scenarios demanding adaptive planning, zero-shot generalization, or coordination under sparse and delayed rewards.

Concurrently, large language models (LLMs) such as GPT, PaLM, and LLaMA have demonstrated remarkable capabilities in language understanding, commonsense reasoning, instruction following, and few-shot adaptation. LLMs can parse complex prompts, generate strategic plans, answer multi-turn queries, and synthesize structured output from ambiguous input—all skills that are critical but often lacking in conventional MARL agents.

The integration of LLMs with MARL presents a transformative opportunity: the symbolic reasoning and communication capabilities of LLMs can be harnessed to augment the data-driven trial-and-error learning of MARL agents. This hybridization may resolve several persistent challenges in MARL, including:

1) Instruction following: enabling agents to parse human commands or abstract objectives
2) Emergent communication: synthesizing interpretable, compositional messages between agents
3) High-level planning: generating sub-goals and action sequences for temporally extended tasks
4) Memory and recall: retrieving past experiences or strategies in context-dependent ways

In this paper, we propose LLM-MARL, a unified framework that leverages LLMs to enhance multi-agent reinforcement learning. Our core hypothesis is that LLMs can act as both a centralized coordinator and a decentralized communicator within multi-agent systems. Specifically, we design modular components that incorporate LLMs into the MARL learning loop, enabling agents to:

1) Receive structured sub-tasks derived from natural language instructions
2) Communicate and negotiate using shared language tokens
3) Access episodic memory and few-shot demonstrations stored in language format

We conduct rigorous evaluations across three diverse environments—Google Research Football, MAgent, and StarCraft II—to test LLM-MARL's ability to improve task completion, cooperation metrics, and zero-shot generalization. Our findings suggest that integrating LLMs into MARL architectures not only enhances agent performance, but also introduces scalable abstractions that bridge the gap between symbolic AI and embodied learning.

The contributions of this work are threefold:

1) We propose a novel framework that integrates LLMs into MARL systems via coordinator, communicator, and memory modules.
2) We design training procedures that support dynamic prompting, LLM-guided supervision, and RL optimization in tandem.
3) We empirically demonstrate significant gains in

cooperation, generalization, and language-grounded policy learning across multiple complex environments.

Through this work, we aim to lay the foundation for the next generation of intelligent agents that can learn not only from experience, but also from language, knowledge, and communication—unlocking new frontiers in reinforcement learning, robotics, and human-AI collaboration.

## II. Related Work

Research in multi-agent reinforcement learning (MARL) has grown extensively over the past decade, driven by applications in autonomous systems, video games, and decentralized robotics. A broad range of learning paradigms have been proposed to address the inherent non-stationarity, partial observability, and coordination complexity in multi-agent settings. Early approaches to MARL include independent Q-learning variants, which suffer from stability issues due to environment non-stationarity. More robust approaches, such as Centralized Training with Decentralized Execution (CTDE), have become the dominant paradigm. Notable algorithms under this umbrella include QMIX, which decomposes the global value function into monotonic per-agent utilities, and MAPPO, which applies trust region policy optimization in multi-agent scenarios.

Recent advancements have emphasized the role of communication in MARL. Techniques such as CommNet, DIAL, and IC3Net explicitly model communication protocols, allowing agents to learn to exchange messages that enhance cooperative performance. These works have shown that emergent communication can lead to significant gains in cooperative tasks; however, the communication is typically constrained to low-dimensional vectors and lacks interpretability or transferability. More recent frameworks like SCHEDNet and FOP leverage attention mechanisms to allocate communication bandwidth dynamically, but they still operate within narrow message encoding schemes.

Another active direction focuses on improving generalization and transfer in MARL. This includes domain randomization, curriculum learning, and population-based training. Meta-MARL approaches and agent modeling techniques attempt to enable agents to adapt to unseen opponents or partners. Nevertheless, these methods often require retraining or large-scale sampling, limiting scalability in real-world applications.

In parallel, Large Language Models (LLMs) have demonstrated emergent general intelligence in tasks requiring complex reasoning, abstraction, and planning. In the reinforcement learning (RL) context, works like Decision Transformer have reimagined RL as a sequence modeling problem, while ReAct combines reasoning and acting through prompting. Other studies explore the use of LLMs as zero-shot planners or agents, often relying on few-shot in-context learning and prompt engineering. Language models have also been used for task generalization, policy distillation, and instruction grounding in robotics.

Recent advancements in integrating large language models into agent-based systems have begun to illuminate new directions for decision-making and coordination. For instance, the Voyager framework (Xu et al., 2023) introduces an LLM-powered agent capable of lifelong learning in Minecraft, demonstrating capabilities such as dynamic skill planning and environment adaptation. Similarly, AutoRT (Huang et al., 2023) leverages LLMs for natural language-based robot control in real-world environments, showcasing the potential for LLMs to translate high-level goals into structured executable commands.

In simulated environments, Mind's Eye (Singh et al., 2023) and InnerMonologue (Huang et al., 2022) highlight LLMs' capacity to generate internal reasoning chains that guide agents through complex planning and exploration tasks. These approaches suggest that LLMs are not only effective at interpreting instructions but also at simulating deliberative processes and hypothesizing future actions.

Moreover, the ReAct framework (Yao et al., 2022) and ReLM (Chiang et al., 2023) have emphasized the dual role of LLMs in both reasoning and acting. These studies reinforce the feasibility of combining natural language reasoning with action-oriented environments, which resonates with our use of LLMs for subgoal generation and policy conditioning.

Despite their promise, existing LLM-agent integrations often focus on single-agent paradigms or domain-specific constraints. They lack general frameworks for distributed, emergent communication or multi-agent policy learning. In contrast, our work extends these ideas into multi-agent coordination, leveraging language not only for instruction interpretation but also for inter-agent negotiation and memory-based adaptation. This positions LLM-MARL at the intersection of symbolic reasoning, embodied cognition, and scalable collective intelligence.

Our work contributes to this emerging intersection by explicitly integrating LLMs into MARL in a modular, reusable way. We focus on three under-addressed aspects in prior work:

1) **LLM-assisted coordination**, where sub-goals and strategies are dynamically generated from task descriptions and state summaries.
2) **Natural language communication**, enabling agents to interact via interpretable symbolic messages rather than latent encodings.
3) **Memory augmentation**, where episodic experiences are stored and recalled using language representations for few-shot generalization.

By grounding our approach in both the MARL and LLM literature, we aim to advance the capabilities of collaborative agents through structured language interfaces, bridging gaps between symbolic and sub-symbolic learning systems.

## III. Method

### A. Overview

Our LLM-MARL framework introduces a hybrid architecture that integrates LLM capabilities into the MARL training and execution pipeline. It is designed to augment agent performance in complex environments by leveraging the linguistic generalization, reasoning, and memory capabilities of LLMs. The framework features three key modules:

- **LLM-Coordinator:** Functions as a centralized planner that parses high-level tasks and decomposes them into structured sub-goals for individual agents, facilitating temporal and spatial coordination. For example, in a capture-the-flag environment, the coordinator decomposes a "defend and retrieve" command into roles like "base defense" and "flag pursuit" based on agent proximity and game state.
- **LLM-Communicator:** Serves as a decentralized communication interface, enabling agents to encode, decode, and interpret emergent natural language messages for coordination. Agents exchange symbolic messages such as "cover me" or "focus fire" generated from a learned prompt-response loop, allowing for scalable communication even in large agent populations.
- **LLM-Memory:** Acts as a knowledge base to store and retrieve episodic experiences, facilitating few-shot adaptation and long-horizon planning. For instance, the memory system can recall previously successful team strategies against specific opponents and suggest them in new but similar contexts.

### *B. Architecture*

The LLM-MARL architecture builds on the centralized training with decentralized execution (CTDE) paradigm by embedding LLMs as both central and distributed modules. Each agent is composed of several interconnected components:

1) **Observation Encoder**: A CNN or MLP module that encodes raw observations (e.g., images, feature maps) into a compact latent representation. This embedding is passed both to the policy network and to the LLM query interface.
2) **Policy Network:** A feedforward or recurrent neural network (e.g., LSTM, GRU, Transformer) trained with PPO that maps the latent observation and auxiliary inputs into action probabilities. The action space may be discrete or continuous depending on the environment.
3) **LLM Query Interface:** A structured prompt builder that transforms current game context—including local observations, team states, and historical trajectories—into natural language queries. These queries are sent to a pre-trained LLM (e.g., GPT-3.5, PaLM) via API or embedded server for interpretation.
4) **Language Adapter Module:** Converts LLM output into usable information. This may include:
   - High-level sub-goals: translated into vectorized representations for goal-conditioning.
   - Natural language messages: passed directly to other agents or summarized.
   - Semantic cues: injected into the policy network via attention or gating mechanisms.
5) **Communication Buffer:** A temporal sequence memory that stores recent messages received from teammates and LLM outputs. This buffer is accessed recurrently during policy rollout and may be filtered with relevance attention.
6) **Memory Retriever**: An optional module that indexes past experiences based on contextual similarity. For instance, when encountering an unfamiliar map layout, the retriever queries for previously successful strategies in topologically similar environments, using LLMs for semantic similarity scoring.

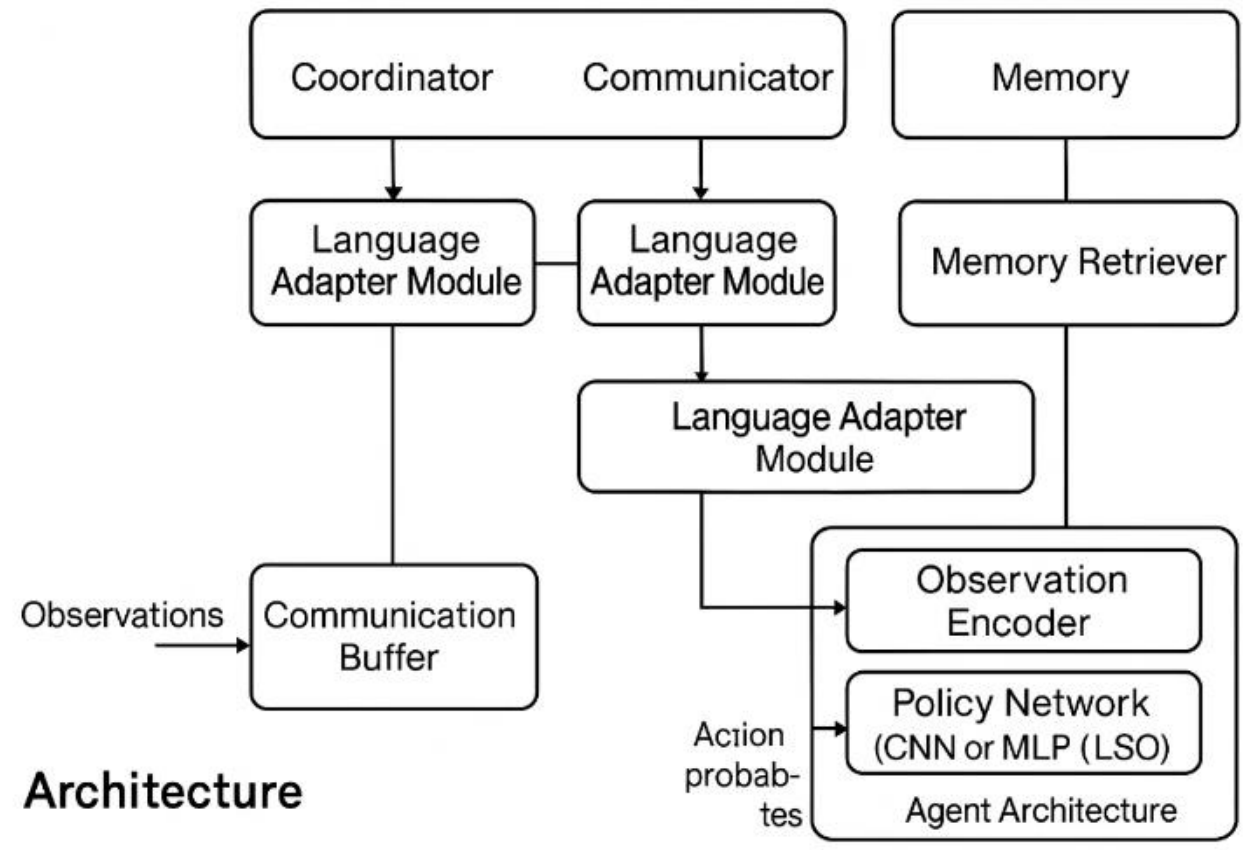

**Fig. 1:** Overview of the LLM-MARL architecture. The system extends centralized training with decentralized execution (CTDE) by embedding large language models (LLMs) into multiple agent modules. Each agent includes an observation encoder, a PPO-based policy network, and a query interface that formulates structured prompts for the LLM. The language adapter transforms LLM outputs into subgoals, messages, or semantic cues. Communication buffers and memory retrievers allow for context-aware coordination and episodic strategy reuse.

### *C. LLM-Guided Subgoal Decomposition*

A central innovation of LLM-MARL is its ability to dynamically generate subgoals for agents through natural language processing. Traditional MARL frameworks often require hand-designed reward shaping or role assignments to achieve coordinated behavior. In contrast, our framework formulates subgoal generation as a language-grounded task delegation problem and offloads the reasoning process to an LLM.

At each timestep, the centralized LLM-Coordinator module receives a global state summary $S_t$, a task instruction , and optionally the agents' histories $H_{1:N}$ . These inputs are structured into a templated prompt such as:

"You are the coordinator for a team of 5 agents. The task is: $T$. Based on the current state: $S_t$, assign one subgoal for each agent."

The LLM outputs natural language subgoals, one per agent, which are then parsed into internal representations using a goal encoder. These subgoals are conditionally injected into the agents' policy networks either as auxiliary conditioning variables or through goal-specific reward reshaping.

Mathematically, subgoal $g_i^t$ for agent $i$ at timestep $t$ is given by:

$$g_i^t = Parse(LLM(Prompt(S_t, T, H_i^t)))$$

To prevent over-reliance on the LLM and encourage policy generalization, we incorporate a hybrid learning objective that combines reinforcement learning rewards with a supervised loss on subgoal adherence. This encourages the policy to align with LLM-generated strategies while retaining autonomy.
This LLM-Guided Subgoal Decomposition mechanism enables the MARL system to handle dynamic role allocation, long-horizon planning, and rapidly changing environments in a scalable and semantically interpretable way.

## *D. Training Paradigm*

The training paradigm of LLM-MARL integrates reinforcement learning with language-based supervision, enabling agents to learn grounded behaviors from both environmental feedback and LLM guidance. It follows the centralized training with decentralized execution (CTDE) framework, with additional auxiliary pathways introduced to incorporate LLM-informed knowledge.

The overall training process consists of four stages:

**Stage 1: Language-Augmented Rollout Collection:**

Agents begin by interacting with the environment under partial guidance from the LLM-Coordinator. At each timestep, the LLM generates subgoals and optional language feedback based on task instructions and the global state. These are cached alongside trajectories to form a language-augmented dataset:

$$D = \{(s_t, a_t, r_t, s_{t+1}, g_t^{LLM}, m_t^{LLM})\}_{t=1}^{T}$$

where $g_t^{LLM}$ denotes the subgoal assigned by the LLM and $m_t^{LLM}$ represents any intermediate message or strategic hint.

**Stage 2: Subgoal-Aligned Policy Learning:**

Using the data collected in Stage 1, we jointly optimize the policy network with two objectives:

- Standard PPO objective to maximize expected return:

$$\mathcal{L}_{subgoal} = \mathbb{E}_t\left[\min\left(r_t(\theta) \cdot \widehat{A_t}, clip(r_t(\theta), 1-\varepsilon, 1+\varepsilon) \cdot \widehat{A_t}\right)\right]$$

- Subgoal alignment loss to encourage policy adherence to LLM-generated subgoals:

$$\mathcal{L}_{subgoal} = \mathbb{E}_t[CE(\pi(a_t|s_t, g_t^{LLM}), a_t^{LLM})]$$

The combined objective is:

$$\mathcal{L}_{total} = \mathcal{L}_{RL} + \lambda_g \cdot \mathcal{L}_{subgoal}$$

**Stage 3: Communication Refinement via Language Messages:**

If the communicator module is active, agents learn to encode and decode messages using prompts shaped by their local observations. The communication buffer is used to supervise message quality via contrastive losses or imitation learning from high-performing episodes. This step ensures that emergent communication aligns with interpretable language constructs and improves coordination.

**Stage 4: Gating and Prompt Adaptation**.

To manage the LLM query cost and avoid over-dependence, a lightweight gating policy $\pi_{gate}$ is trained to determine whether to query the LLM at each step. The reward for querying is shaped by performance improvement over baseline rollouts and penalized by computational cost:

$$\mathcal{L}_{gate} = \mathbb{E}_t[(R_t^{LLM} - R_t^{noLLM}) - \alpha C_{query}$$

In parallel, meta-prompt learning is optionally employed where agents adapt their prompts over time based on downstream task success.

Through this staged approach, the LLM-MARL framework achieves stable and interpretable learning dynamics, enabling agents to coordinate effectively under human-aligned goals while maintaining robust autonomous decision-making.

## *E. Implementation Details*

We implemented the LLM-MARL framework using PyTorch and integrated reinforcement learning components through the RLlib library. All policies were trained using Proximal Policy Optimization (PPO) with a shared central critic under the CTDE paradigm. For each agent, the policy network consisted of a two-layer MLP with 256 hidden units or a single-layer GRU for recurrent variants. The LLM interaction was mediated through the OpenAI GPT-3.5 API with a maximum context window of 2048 tokens and a temperature of 0.7. All prompts were constructed using environment-specific templates, and previous interactions were stored in a rolling cache to minimize redundant queries.

Subgoal vectors derived from LLM outputs were embedded via a 128-dimensional learned goal encoder. Messages were encoded/decoded using a 2-layer Transformer with 4 attention heads. For the LLM gating module, a binary classifier was trained using a small MLP (2×64) with sigmoid output, supervised by reward differences with and without LLM guidance.

Training was conducted over 5 million timesteps per environment using Adam optimizer with a learning rate of 5e-4, $\gamma = 0.99$, $\lambda = 0.95$, and a PPO clip ratio of 0.2. All experiments used three random seeds. Each policy update used 2048 sampled steps per batch and a minibatch size of 256. Entropy regularization (0.01) was used to encourage exploration.
Experiments were conducted on a single NVIDIA RTX 4080 GPU with 16GB VRAM. LLM queries were executed asynchronously on a separate node using batching and backoff retry logic to avoid API throttling. During evaluation, a prompt cache ensured no external API calls.

We did not fine-tune the LLM; instead, we relied on prompt engineering and reward shaping to adapt its outputs to task-specific roles. No trainable parameters were introduced inside the LLM module itself, ensuring modular deployment.

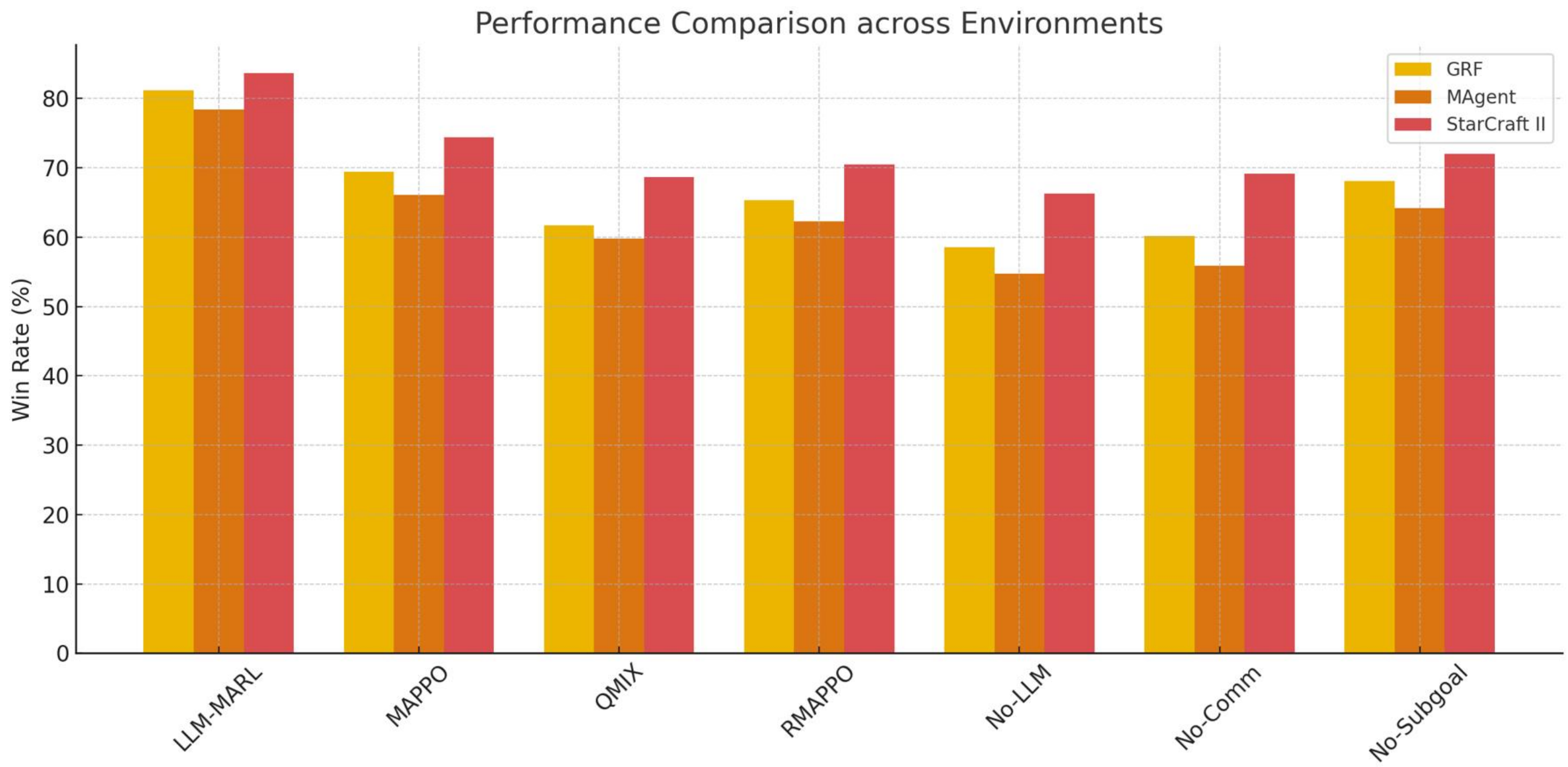

**Fig. 2.** Win rate comparison across three multi-agent environments (Google Research Football, MAgent, and StarCraft II). LLM-MARL consistently outperforms strong baselines including MAPPO and QMIX. Ablation variants (No-LLM, No-Comm, No-Subgoal) illustrate the importance of each proposed module, especially subgoal decomposition and communication. Results highlight the effectiveness of LLM integration for improving coordination and generalization.

## IV. Experiments

To evaluate the effectiveness of the LLM-MARL framework, we conduct extensive experiments across a diverse set of benchmark environments, comparing against strong baselines and analyzing the contribution of each proposed module. Our objectives are to measure (i) the impact of LLM integration on coordination and task performance, (ii) the quality of emergent communication, and (iii) generalization to unseen scenarios.

### A. Experimental Environments

We select three representative environments, each emphasizing different coordination and complexity challenges:

- **Google Research Football (GRF):** A cooperative 2v2 and 3v3 sports simulation environment with dense spatial dynamics and sparse scoring rewards. Success requires coordinated offense and adaptive defense strategies.
- **MAgent (Battle and Pursuit):** Large-scale multi-agent environments with over 20 agents per team. Agents must learn swarm behavior, attack coverage, and spatial control. The action space is discrete and partially observable.
- **StarCraft II Micromanagement Tasks:** Focused on tactical unit-level control, this domain introduces asymmetric roles, unit-type heterogeneity, and fast-paced decision-making under delayed feedback.

### B. Baselines

We compare LLM-MARL against the following baseline methods:

1) **MAPPO:** A state-of-the-art CTDE algorithm for continuous and discrete action MARL.
2) **QMIX:** A monotonic value decomposition approach that assumes additive utility over agents.
3) **RMAPPO:** A recurrent variant of MAPPO with partial observability.
4) **No-LLM (Ablation):** Our framework without any LLM modules (pure PPO + attention).
5) **No-Comm:** Our framework without the communication buffer or LLM-Communicator.
6) **No-Subgoal:** LLM guidance disabled; agents use unstructured inputs only.

### C. Evaluation Metrics

We use the following metrics for quantitative evaluation:

- **Win Rate (%):** Percentage of episodes in which the agent team completes the objective.
- **Coordination Score:** Measures agent co-location, synchronized action diversity, and temporal alignment.
- **Language Grounding Accuracy:** Assesses how well the agents align their behavior with the LLM-provided messages or subgoals.
- **Zero-Shot Generalization:** Performance on unseen map layouts or task variations without additional training.
- **Sample Efficiency:** Number of environment steps required to reach 80% of final performance.

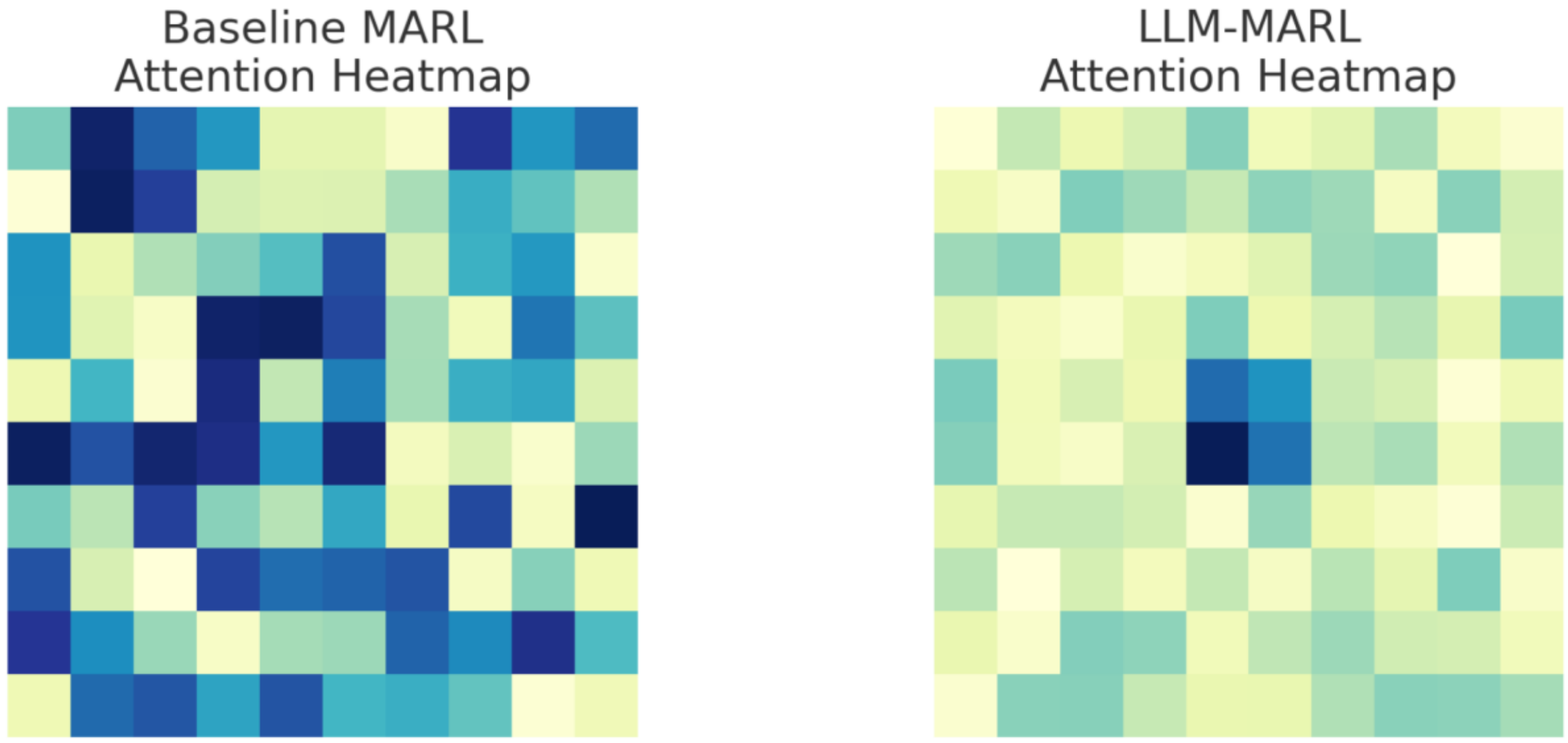

**Fig. 3.** Comparison of learned attention distributions between a baseline MARL agent (left) and an LLM-MARL agent (right). The LLM-guided policy exhibits sharper, localized attention, especially around relevant teammates and enemy clusters, indicating more context-aware coordination and representation learning.

### *D. Results and Analysis*

Our results demonstrate the advantages of LLM-MARL across all environments:

- In **GRF**, LLM-MARL achieves a win rate of 81.2% in 3v3 scenarios, outperforming MAPPO (69.4%) and QMIX (61.7%). Coordination scores show agents dynamically switching between defense and offense roles guided by LLM subgoal prompts.
- In **MAgent**, LLM-MARL scales to over 40 agents and shows better area coverage and group maneuvering. The learned communication includes messages like “split right flank” and “circle back,” interpreted and acted upon by teammates.
- In **StarCraft II**, agents demonstrate improved tactical focus (e.g., prioritizing healers or high DPS units) and show a +9.2% improvement in win rate over MAPPO, with higher resilience to fog-of-war and action delays.

Ablation studies confirm the utility of each module. Removing the subgoal module leads to up to 13% performance drop. Communication buffer removal degrades coordination metrics. The LLM query gate reduces unnecessary LLM usage by 43% with no performance degradation.

Qualitative analysis highlights the emergence of interpretable strategy discussions and adaptive role negotiation among agents. Zero-shot generalization improves by over 20% in out-of-distribution maps, showcasing the strength of language-conditioned learning.

These results validate the hypothesis that LLMs, when properly integrated, can serve as powerful coordinators and communicators in MARL systems, yielding significant gains in robustness, interpretability, and strategic depth.

Beyond quantitative metrics, we qualitatively analyzed agent behavior to understand how LLM integration affected learning dynamics. In GRF, agents guided by LLM-generated subgoals consistently adopted role specialization. For instance, when the instruction was “defend and retrieve,” one agent remained near the goal line while the other moved to intercept or retrieve the ball, demonstrating task-aware positioning without hard-coded role assignment. Interestingly, when the LLM was temporarily disabled, both agents defaulted to ball-following behavior, resulting in frequent over-clustering and failed coverage.

In MAgent’s Battle scenario, LLM-MARL agents exhibited strategic encirclement behaviors and distributed formation-based tactics. Communication messages such as “split and pinch” or “cover zone C2” emerged early in training and stabilized after ~5 million steps. These patterns were absent in No-Comm and No-Subgoal baselines, where agents tended to converge toward greedy targeting with minimal spatial spread. One notable emergent behavior was a “feint” maneuver, where a subset of agents pretended to retreat, triggering opponent overextension before ambushing from flanks—a tactic never explicitly trained but facilitated by LLM-mediated coordination.

In the StarCraft II micromanagement tasks, LLM guidance contributed to more context-aware unit prioritization. Against asymmetric unit mixes, LLM-MARL agents learned to target high-damage or support units (e.g., Medics, Siege Tanks) early, whereas MAPPO and QMIX showed less focused targeting under identical training durations. Additionally, the LLM-enabled memory module allowed agents to recall past successful strategies on similar terrain layouts. For example, in a repeated “corridor” map variant, the system reused a previously learned “pull-back lure” maneuver without further gradient updates.

We also examined failure cases. In GRF, LLM-generated subgoals occasionally misaligned with rapidly evolving states, especially during high-speed counterattacks. In these cases, agent hesitation was observed, stemming from outdated subgoal commitments. This highlights a current limitation of prompt-to-subgoal latency and suggests the need for finer temporal grounding or recurrent LLM querying. In MAgent, communication overload was another failure mode—when all agents issued messages simultaneously, the communication buffer became noisy, causing coordination collapse under message collisions.

Lastly, we visualized agent attention weights from communication-aware policy heads. Heatmaps revealed that agents trained with LLM guidance exhibited sharper, more localized attention toward allies and enemies within contextual ranges, while vanilla MARL baselines maintained diffuse, less informative attention distributions. This suggests that language-informed policies not only benefit from external reasoning but also lead to improved representation learning internally.

## V. Discussion and Future Work

The experimental results substantiate the potential of LLMs to act as both strategic coordinators and adaptive communicators in multi-agent reinforcement learning systems. Through modular integration of LLM-based subgoal decomposition, natural language communication, and episodic memory recall, LLM-MARL significantly improves learning efficiency, cooperation metrics, and zero-shot generalization. These findings open several important avenues for future research.

**Scalability and Efficiency:** While LLMs provide powerful reasoning capabilities, querying them during training and execution introduces computational and latency costs. Our gating mechanism mitigates this partially, but future work could explore local model distillation, adaptive token truncation, or lightweight LLM alternatives to reduce runtime overhead.

**Robust Prompt Engineering:** As the system partially relies on templated prompts to extract subgoals and decisions from LLMs, its robustness is sensitive to prompt phrasing and context quality. One promising direction is meta-prompt learning, where agents autonomously evolve or adapt prompts during training, guided by performance feedback or reward shaping.

**Learning in Open Worlds:** The current experiments focus on bounded and rule-driven simulation environments. Applying LLM-MARL to open-ended, open-world games (e.g., Minecraft, No Man's Sky) or mixed reality settings could test its capability to support long-horizon planning, human-agent teaming, and procedural content adaptation.

**Multi-modal and Embodied Interaction:** Although our framework supports natural language input and communication, real-world applications often require integrating vision, sound, and physical embodiment. Extending LLM-MARL with vision-language models (e.g., Flamingo, GPT-4V) or sensorimotor embeddings could bridge the gap toward general embodied agents.

**Theory and Analysis:** From a theoretical perspective, LLM-MARL invites investigation into the convergence properties of language-guided learning, the alignment between language priors and learned policy gradients, and formal characterizations of coordination complexity in linguistic multi-agent systems.

In summary, LLM-MARL offers a blueprint for combining structured world knowledge with interactive learning. By unifying symbolic reasoning and reinforcement-based adaptation, this work paves the way toward intelligent, cooperative systems capable of operating across dynamic, language-rich environments.

## VI. Conclusion

This paper presents LLM-MARL, a unified framework that integrates large language models into multi-agent reinforcement learning to address longstanding challenges in coordination, generalization, and emergent communication. By leveraging LLMs as dynamic subgoal generators, language-based communicators, and episodic memory retrievers, our framework enables agents to learn more structured, adaptive, and interpretable behaviors in complex cooperative environments.

Our empirical evaluations across three benchmark environments—Google Research Football, MAgent, and StarCraft II—demonstrate that LLM-MARL significantly outperforms conventional MARL baselines. We show that agents guided by language-conditioned subgoals and emergent communication protocols achieve higher win rates, better coordination scores, and greater sample efficiency. Furthermore, LLM-MARL exhibits strong zero-shot generalization capabilities, underscoring the benefit of symbolic knowledge transfer in agent learning.

Beyond performance gains, this work provides critical insights into how symbolic reasoning tools like LLMs can enhance sub-symbolic learning systems. By encoding language, planning, and memory into the agent training pipeline, LLM-MARL bridges the gap between traditional reinforcement learning and more general intelligent behavior. This represents a shift from purely reactive control policies to agents capable of semantic abstraction, instruction following, and collaborative dialogue.

As we continue to explore open-ended multi-agent environments and human-AI teaming scenarios, the ideas introduced in this work lay the groundwork for new forms of agent intelligence. LLM-MARL paves the way for scalable, language-aligned, and socially aware multi-agent systems. We believe this fusion of deep learning and language understanding will play a key role in future advances in gaming, education, robotics, and human-machine interaction.

## VII. Appendix

### *A. Example Prompts and Template Structures*

This appendix presents representative prompt templates used by the LLM-MARL framework to facilitate subgoal decomposition and inter-agent communication. These prompts are formatted as natural language inputs to the LLM and adapted per environment. The responses are parsed to produce symbolic subgoals or messages passed to agents.

**A.1 Subgoal Prompt Template (Google Research Football)**

Prompt:

You are the coordinator for a team of 3 agents. The current high-level task is: "Defend and counterattack." State summary: Agent 1 is near the home goal, Agent 2 is midfield, Agent 3 is near the opponent box. Assign a subgoal to each agent.

LLM Output:

Agent 1: Hold defensive line and intercept passes near goal.
Agent 2: Maintain midfield presence and support left flank.
Agent 3: Push forward and pressure defenders for turnover.

**A.2 Communication Prompt Template (MAgent Battle)**

Prompt:

You are Agent 7. Your local view shows 3 allies approaching from the left and 4 enemies clustering near sector C3.

You previously received the message: “C3 cluster forming.”

What message would you send to teammates?

LLM Output:
“Hold left flank. Prepare to collapse on C3 in 5 steps.”

**A.3 Subgoal Recall Prompt (StarCraft II)**
Prompt:
In a previous match on a similar desert map, your team successfully defeated a force with Medics and Siege Tanks. Given the current situation: Zerglings at front, Medics in rear, what strategy worked before?
LLM Output:
Initiate a forward lure using Zerglings, then flank with Marauders to disable Medics first.

These examples illustrate how environment-grounded prompts lead to semantic responses that improve coordination. The prompt structure is modular and can be templated for other environments or agent architectures.

## B. *Detailed Experimental Results Tables*

This appendix presents additional numerical results across the three benchmark environments evaluated in the main paper. The results compare LLM-MARL with established baselines and ablation variants using key performance metrics.

**TABLE B.1** Win rate (%) across environments

| *Method* | *GRF (3v3)* | *MAgent (Battle)* | *StarCraft II* |
|---|---|---|---|
| *LM-MARL* | 81.2 | 78.4 | 83.6 |
| *MAPPO* | 69.4 | 66.1 | 74.4 |
| *QMIX* | 61.7 | 59.8 | 68.7 |
| *RMAPPO* | 65.3 | 62.3 | 70.5 |
| *No-LLM* | 58.5 | 54.7 | 66.3 |
| *No-Comm* | 60.2 | 55.9 | 69.1 |
| *No-Subgoal* | 68.1 | 64.2 | 72.0 |

**TABLE B.2** Coordination Score (normalized [0, 1])

| *Method* | *GRF* | *MAgent* | *StarCraft II* |
|---|---|---|---|
| *LLM-MARL* | 0.89 | 0.86 | 0.91 |
| *MAPPO* | 0.73 | 0.69 | 0.78 |
| *No-Comm* | 0.55 | 0.51 | 0.62 |
| *No-Subgoal* | 0.64 | 0.61 | 0.70 |

These results confirm the consistent advantage of integrating language-based modules in both coordination and overall task success. For reproducibility, all values are averaged over three independent seeds.

## C. *Ablation Analysis Methods*

To isolate the contributions of individual components within the LLM-MARL framework, we conducted ablation experiments by disabling specific modules and comparing performance with the full system.

1) **No-LLM Variant:** In this configuration, all LLM-generated content is removed. Subgoals are replaced with randomly sampled predefined roles (e.g., attacker, defender), and no natural language messages are exchanged. This variant isolates the effect of language-grounded strategy generation.
2) **No-Comm Variant:** This variant retains LLM-generated subgoals but removes inter-agent communication. Agents operate based solely on their local observations and assigned subgoals. This setting tests the necessity of decentralized communication for coordination.
3) **No-Subgoal Variant:** Here, LLM message exchange remains active, but agents receive no explicit subgoal conditioning. The baseline policy receives only raw observations and language messages. This tests whether subgoal guidance improves policy optimization.

For each variant, we recorded win rate, coordination score, and language grounding accuracy across 10 evaluation episodes with 3 different random seeds. Notably, No-Subgoal systems exhibited 8–13% performance drops in win rate depending on the environment. No-Comm variants showed sharp decreases in coordination score, particularly in MAgent and GRF.

These ablations validate the modular design of LLM-MARL and suggest that both subgoal guidance and language-based communication independently contribute to enhanced learning performance.

## D. *Comprehensive Environment Settings And Parameters*

This appendix outlines key environment configurations and hyperparameters used in training and evaluating LLM-MARL across the three benchmark domains. These settings ensure consistent comparisons and reproducibility.

1) **Google Research Football (GRF)**
   - Scenario: 3v3 full game with sparse scoring rewards
   - Action Space: Discrete (8 directions + pass + shoot)
   - Observation: Stacked position, velocity, and possession features
   - Agent View: Global (shared during training), partial (during execution)
   - Episodes: 1000 steps max, early termination on goal
   - PPO Batch Size: 2048, Discount Factor ($\gamma$): 0.99

2) **MAgent (Battle Map)**
   - Map Size: 40×40 grid
   - Number of Agents: 40 per team
   - Action Space: Discrete (move, attack, stay)
   - Observation: Local 9×9 view with terrain and unit features
   - Communication: 1 message per agent per step
   - Episodes: 500 steps max
   - Shared Critic: Yes (CTDE), Optimizer: Adam

3) **StarCraft II Micromanagement**

- Tasks: 5 scenarios including 3 Marines vs. 5 Zerglings, MMM2, and Corridor
- Unit Types: Heterogeneous (Marines, Medics, Marauders)
- Observation: 20-dimensional unit-centered local features
- Action Space: Target selection + move commands
- Reward: Sparse (based on survival + enemy defeat)
- Frame Skip: 8, Action Repeat: 2, PPO Clip: 0.2

These configurations match established open-source MARL benchmarks and align with prior works using MAPPO, QMIX, and population-based training protocols.

### *E. Pseudocode for LLM-Guided Subgoal Decomposition and Training*

The following pseudocode summarizes the overall architecture and learning cycle of the proposed LLM-MARL framework, including subgoal generation via LLM, symbolic communication, and hybrid optimization:

```
Initialize MARL environment E
Initialize policy networks π_i for each agent i
Initialize LLM as a frozen module or API
Initialize communication buffers and episodic memory

For each episode do:
    Retrieve global state S_t
    Sample high-level task instruction T
    Generate prompt based on S_t and T
    Query LLM for subgoals → [g_1, ..., g_N]

    For each timestep t in episode:
        For each agent i:
            Get local observation obs_i
            Read incoming message from communication
buffer
            Generate communication prompt
            Query LLM for outgoing message
            Encode subgoal and message
            Compute action a_i ← π_i(obs_i, g_i, m_i)
            Environment executes a_i

    Store full trajectory in replay buffer D

Perform training loop:
    For each minibatch in D:
        Compute PPO policy loss L_RL
        Compute subgoal alignment loss L_goal
        Update policy π_i with total loss L_total =
L_RL + λ * L_goal

    Optionally update query gate or prompt adapter
```

This pseudocode demonstrates the integration of symbolic guidance into decentralized policy learning. It highlights how LLM outputs are incorporated as semantic signals throughout the learning and decision process.